\documentclass[11pt]{article}

\usepackage[final]{acl}

\usepackage{times}
\usepackage{latexsym}

\usepackage[T1]{fontenc}

\usepackage[utf8]{inputenc}

\usepackage{microtype}

\usepackage{graphicx}
\usepackage{amsmath}
\usepackage{amssymb}
\usepackage{amsfonts}
\usepackage{multirow}
\usepackage{booktabs}
\usepackage{enumitem}
\usepackage{float}

\title{Memory-Efficient Structured Backpropagation for \\On-Device LLM Fine-Tuning}

\author{
Juneyoung Park, Yuri Hong, Seongwan Kim{$^\dagger$}, Jaeho Lee{$^\dagger$}
\\ \\ \textbf{OptAI Inc.} \\
 \texttt{\{jyoung.park, yri.hong, swan.kim, jaeho.lee\}@opt-ai.kr}
}

\begin{document}
\maketitle

\renewcommand{\thefootnote}{\fnsymbol{footnote}}
\setcounter{footnote}{2}
\footnotetext{Corresponding Author}

\begin{abstract}
On-device fine-tuning enables privacy-preserving personalization of large language models, but mobile devices impose severe memory constraints, typically 6--12GB shared across all workloads. Existing approaches force a trade-off between exact gradients with high memory (MeBP) and low memory with noisy estimates (MeZO). We propose Memory-efficient Structured Backpropagation (MeSP), which bridges this gap by manually deriving backward passes that exploit LoRA's low-rank structure. Our key insight is that the intermediate projection $h = xA$ can be recomputed during backward at minimal cost since rank $r \ll d_{in}$, eliminating the need to store it. MeSP achieves 49\% average memory reduction compared to MeBP on Qwen2.5 models (0.5B--3B) while computing mathematically identical gradients. Our analysis also reveals that MeZO's gradient estimates show near-zero correlation with true gradients (cosine similarity $\approx$0.001), explaining its slow convergence. MeSP reduces peak memory from 361MB to 136MB for Qwen2.5-0.5B, enabling fine-tuning scenarios previously infeasible on memory-constrained devices. We provide an example implementation of our proposed method at \url{https://github.com/crinex/acl-mesp}
\end{abstract}

\section{Introduction}
On-device fine-tuning of large language models has emerged as a promising approach for privacy-preserving personalization, enabling users to adapt models to their data without transmitting sensitive information to cloud servers. However, this capability remains largely theoretical for most mobile devices. Even with parameter-efficient methods like LoRA~\cite{hu2021lora}, the memory footprint of intermediate activations during backpropagation exceeds the 6--12GB typically available on mobile devices. For example, fine-tuning a modest 0.5B parameter model at sequence length 256 requires over 360MB of peak memory under standard memory-efficient backpropagation, a significant burden when this memory must be shared with the operating system and other applications.

Existing approaches force practitioners into an unsatisfying trade-off. Zeroth-order methods like MeZO~\cite{malladi2023mezo} achieve inference-level memory usage by estimating gradients through forward passes alone, but their gradient estimates exhibit variance proportional to parameter count, requiring 10--100$\times$ more iterations to converge. Memory-efficient backpropagation (MeBP)~\cite{apple2024mebp} computes exact gradients with gradient checkpointing, but delegates tensor lifecycle decisions to automatic differentiation frameworks, which store more intermediates than mathematically necessary.

Interestingly, we observe that LoRA's low-rank structure offers an unexploited opportunity for memory optimization. The intermediate projection $h = xA$ has shape $[\text{batch}, \text{seq}, r]$ where the rank $r$ is typically 8--32. Since $r \ll d_{in}$, recomputing $h$ during the backward pass costs far less than storing it across all LoRA layers. Existing methods treat this computation as a black box, but by manually deriving backward passes, we can determine precisely which tensors to store, when to release them, and what to recompute.

Can we achieve the memory efficiency of zeroth-order methods while preserving the gradient accuracy of first-order backpropagation? In this paper, we answer affirmatively with Memory-efficient Structured Backpropagation (MeSP). Our approach processes transformer blocks~\cite{vaswani2017attention} sequentially, storing only block outputs during forward and recomputing all intermediates during backward. By deriving explicit backward passes for each transformer component, MeSP provides fine-grained control over tensor lifecycles that automatic differentiation cannot achieve. As we demonstrate in Section~\ref{sec:experiments}, this reduces peak memory by 62\% for Qwen2.5-0.5B while computing mathematically identical gradients.

Our contributions are as follows:
\begin{itemize}[itemsep=0pt, parsep=0pt, topsep=0pt]
    \item We propose MeSP, a memory-efficient training algorithm that achieves 49\% average memory reduction compared to MeBP while preserving first-order gradient accuracy through manually derived backward passes.
    \item We demonstrate that MeSP reduces peak memory from 361MB to 136MB (62\% reduction) for Qwen2.5-0.5B on Apple Silicon, with only 28\% computational overhead.
    \item We reveal that MeZO's gradient estimates are essentially uncorrelated with true gradients (cosine similarity $\approx$0.001), providing new insight into why zeroth-order methods converge slowly.
    \item We release our MLX-based implementation for reproducible on-device training research.
\end{itemize}

\section{Related Work}
\begin{figure*}[t]
\centering
\includegraphics[width=\textwidth]{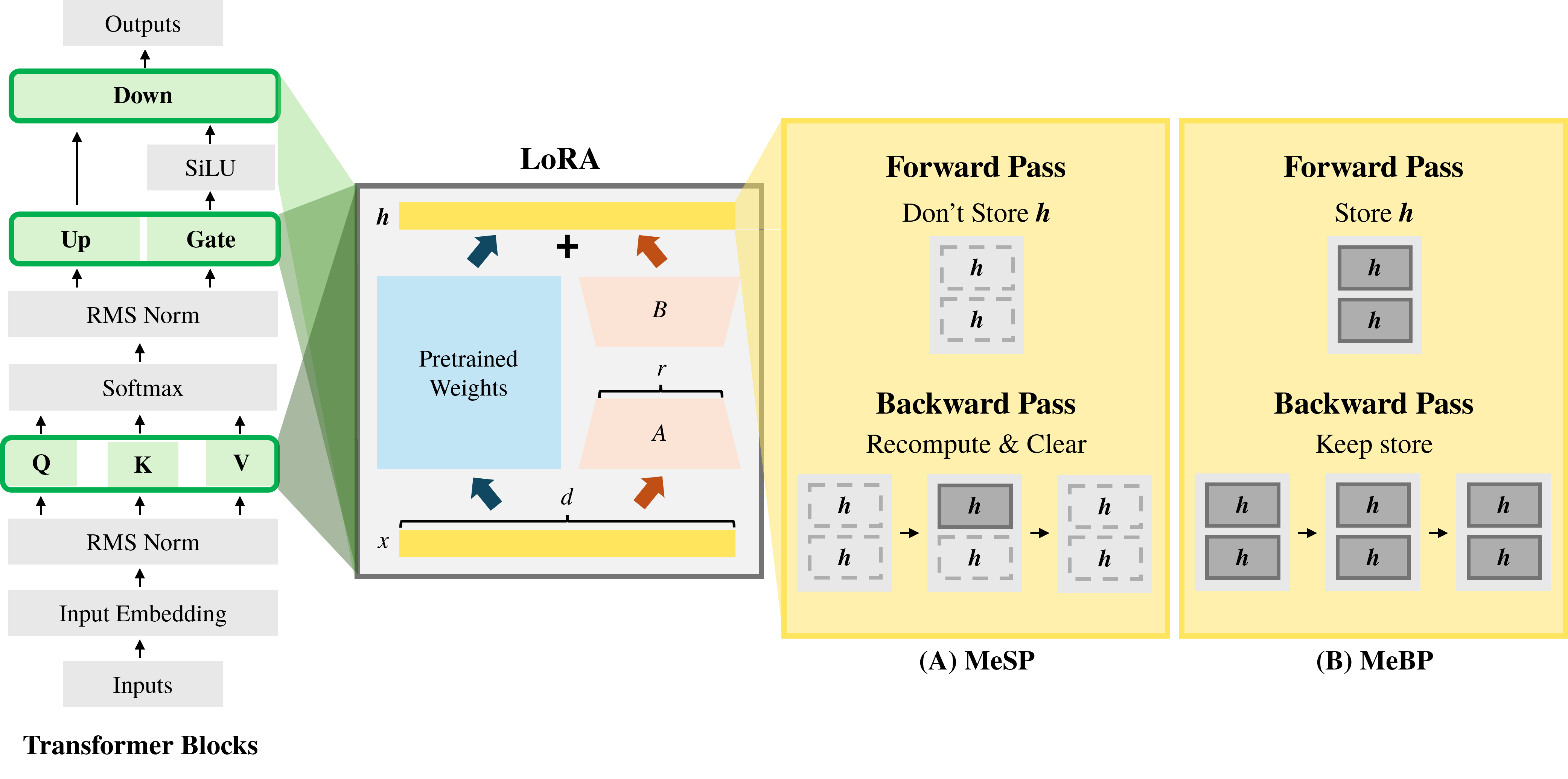}
\caption{Visualization of LoRA fine-tuning with Memory-efficient Structured Backpropagation (MeSP) compared to Memory-efficient Backpropagation (MeBP). (A) In MeSP, the intermediate LoRA projection $h = xA$ is not cached in the forward pass and is recomputed in the backward pass, then discarded immediately after being used to compute the gradients of $A$ and $B$. This keeps only a few essential activations in memory at any time, substantially reducing peak memory usage. (B) In MeBP, the intermediate projection $h$ is kept as a forward activation for each LoRA module and later loaded again during backpropagation to compute the gradients. Because these $h$ tensors remain in memory, this leads to higher peak memory usage than MeSP.}
\label{fig:overview}
\end{figure*}

\paragraph{Memory-Efficient Training.}
Gradient checkpointing~\cite{chen2016training} reduces memory by recomputing activations during backward, achieving $O(\sqrt{n})$ memory complexity. FlashAttention~\cite{dao2022flashattention,dao2023flashattention2} applies this principle to attention, recomputing softmax weights rather than storing the full attention matrix. MeZO~\cite{malladi2023mezo} takes a more radical approach, eliminating backpropagation entirely through zeroth-order gradient estimation. However, the variance of these estimates scales with parameter count, degrading convergence for large models. MeBP~\cite{apple2024mebp} combines gradient checkpointing with LoRA on Apple Silicon, but relies on automatic differentiation which implicitly retains more tensors than necessary. \textit{In contrast, our work provides explicit control over tensor lifecycles by manually deriving backward passes, achieving memory efficiency comparable to zeroth-order methods while preserving first-order gradient accuracy.}

\paragraph{Parameter-Efficient Fine-Tuning.}
Adapter methods~\cite{houlsby2019parameter} insert trainable bottleneck modules between frozen layers. LoRA~\cite{hu2021lora} parameterizes weight updates as low-rank matrix products $\Delta W = AB$, reducing trainable parameters from $O(dk)$ to $O(r(d+k))$ where $r \ll d, k$. QLoRA~\cite{dettmers2023qlora} combines LoRA with 4-bit quantization, achieving up to $4\times$ memory reduction for model weights. Despite these advances in reducing parameter count, the memory footprint of intermediate activations during training remains a bottleneck for on-device deployment. \textit{Unlike previous approaches that focus solely on parameter efficiency, we exploit LoRA's low-rank structure to optimize activation memory by recomputing the small intermediate projection $h = xA$ rather than storing it.}

\paragraph{On-Device Training.}
Lin et al.~\cite{lin2022tiny} demonstrated training under 256KB memory through aggressive quantization and sparse updates. PockEngine~\cite{xu2023pockengine} optimizes computational graphs for mobile deployment. Apple's MLX~\cite{apple2023mlx} provides native machine learning support on Apple Silicon with unified memory architecture. \textit{We build upon this ecosystem, contributing memory-efficient training algorithms specifically designed to exploit the unique properties of LoRA fine-tuning on unified memory devices.}

\section{Background}
This section provides the technical foundation for our approach, covering Low-Rank Adaptation, zeroth-order optimization, and memory-efficient backpropagation.

\subsection{Low-Rank Adaptation (LoRA)}

Low-Rank Adaptation~\cite{hu2021lora} is a parameter-efficient fine-tuning method that freezes pre-trained weights and introduces trainable low-rank decomposition matrices. For a linear layer with frozen weight
$W_0 \in \mathbb{R}^{d_{in} \times d_{out}}$, the forward computation becomes
\begin{equation}
y = x W_0 + s \cdot x A B ,
\end{equation}
where $A \in \mathbb{R}^{d_{in} \times r}$ and $B \in \mathbb{R}^{r \times d_{out}}$ are trainable matrices, $r \ll \min(d_{in}, d_{out})$ is the rank, and $s = \alpha / r$ is a scaling factor. This formulation reduces the number of trainable parameters from $d_{in} \times d_{out}$ to $r \times (d_{in} + d_{out})$.

Given the upstream gradient $\partial L / \partial y$, the gradients for the LoRA parameters are computed as
\begin{equation}
\frac{\partial L}{\partial B}
= s \cdot h^{\top} \frac{\partial L}{\partial y},
\qquad
\frac{\partial L}{\partial A}
= x^{\top} \frac{\partial L}{\partial h},
\end{equation}
where $h = x A$ is the intermediate projection and
\begin{equation}
\frac{\partial L}{\partial h}
= s \cdot \frac{\partial L}{\partial y} B^{\top}.
\end{equation}
A key observation is that $h$ has shape $[\text{batch}, \text{seq}, r]$. Since $r$ is typically small, in the range of 8 to 32, recomputing $h$ during the backward pass requires only
$O(\text{batch} \times \text{seq} \times d_{in} \times r)$ operations. This cost is substantially lower than the memory access overhead incurred by storing and retrieving $h$ for long sequences.

\subsection{Zeroth-Order Optimization (MeZO)}

Zeroth-order optimization estimates gradients without backpropagation by using finite differences. MeZO~\cite{malladi2023mezo} approximates the gradient as
\begin{equation}
\nabla L(w) \approx
\frac{L(w + \epsilon z) - L(w - \epsilon z)}{2 \epsilon} \cdot z,
\end{equation}
where $z \sim \mathcal{N}(0, I)$ is a random perturbation vector and $\epsilon$ is a small constant. The method requires two forward passes per update, one with positively perturbed parameters and one with negatively perturbed parameters. The resulting gradient estimate is then used to update the weights.

This approach achieves memory consumption equivalent to inference, as no intermediate activations need to be stored for backpropagation. However, the variance of the gradient estimate scales linearly with the parameter dimensionality $d$, that is,
$\mathrm{Var}[\hat{g}] = O(d)$. For models with billions of parameters, this high variance severely degrades the signal-to-noise ratio of the gradient estimates and typically requires 10 to 100 times more iterations to achieve loss reduction comparable to first-order methods~\cite{malladi2023mezo}.

\subsection{Memory-Efficient Backpropagation (MeBP)}

Memory-efficient backpropagation combines gradient checkpointing with automatic differentiation to reduce peak memory usage while computing exact gradients. During the forward pass, only layer outputs are stored as checkpoints, while intermediate activations within each layer are discarded. During the backward pass, the forward computation for each layer is re-executed to regenerate the required intermediates before computing gradients.

This approach reduces memory complexity from $O(L \times I)$ to $O(L \times O + I)$, where $L$ denotes the number of layers, $I$ denotes the size of intermediate activations per layer, and $O$ denotes the output size per layer. Since $O \ll I$ for typical transformer architectures, this leads to substantial memory savings.

However, MeBP relies on automatic differentiation frameworks such as \texttt{mx.grad} in MLX or \texttt{torch.autograd} in PyTorch to compute gradients within each checkpointed segment. These frameworks treat the computation as a black box and implicitly determine which tensors to retain for gradient computation. As a result, more intermediate tensors may be stored than are mathematically necessary, since the framework cannot exploit domain-specific structure such as the low-rank property of LoRA projections.




\section{Method}
\subsection{Core Idea: Trading Computation for Memory}

The fundamental insight behind MeSP is simple: \textit{small tensors are cheaper to recompute than to store}. In LoRA, the intermediate projection $h = xA$ has dimensions $[\text{batch}, \text{seq}, r]$ where the rank $r$ is typically 8--32. Storing $h$ across all LoRA layers (7 per transformer block $\times$ 24 blocks = 168 layers for Qwen2.5-0.5B) accumulates significant memory. However, recomputing $h$ requires only a single matrix multiplication with the small matrix $A$, which is fast compared to the memory access overhead of storage and retrieval.

This observation leads to our strategy: during the forward pass, we store only the outputs of each transformer block, just enough to restart computation during backward. During the backward pass, we recompute all intermediates on-demand, process one layer at a time, and immediately release memory after computing gradients. At any point, only a single layer's intermediates reside in memory.

\begin{table}[t]
\centering
\small
\begin{tabular}{@{}llrrr@{}}
\toprule
\textbf{Model} & \textbf{Method} & \textbf{Mem (MB)} & \textbf{Time (s)} & \textbf{Red.} \\
\midrule
\multirow{3}{*}{0.5B}
 & MeBP & 360.8 & 0.68 & -- \\
 & MeZO & 243.0 & 0.51 & 33\% \\
 & MeSP & \textbf{136.2} & 0.86 & \textbf{62\%} \\
\midrule
\multirow{3}{*}{1.5B}
 & MeBP & 516.2 & 1.66 & -- \\
 & MeZO & 376.0 & 1.21 & 27\% \\
 & MeSP & \textbf{262.6} & 2.17 & \textbf{49\%} \\
\midrule
\multirow{3}{*}{3B}
 & MeBP & 637.6 & 3.21 & -- \\
 & MeZO & 479.2 & 2.24 & 25\% \\
 & MeSP & \textbf{368.4} & 4.09 & \textbf{42\%} \\
\bottomrule
\end{tabular}
\caption{Memory and time comparison at sequence length 256. MeSP achieves 42--62\% memory reduction vs MeBP while computing exact gradients.}
\label{tab:main_results}
\end{table}

\subsection{Explicit Gradient Computation for LoRA}

We now formalize how gradients are computed without storing the intermediate $h$. Consider a LoRA layer with frozen weight $W_0$, trainable matrices $A$ and $B$, and scaling factor $s = \alpha/r$:
\begin{equation}
y = xW_0 + s \cdot xAB
\end{equation}
Given the upstream gradient $g = \partial L / \partial y$, the parameter gradients can be expressed as:
\begin{equation}
\frac{\partial L}{\partial B} = h^\top (s \cdot g), \qquad \frac{\partial L}{\partial A} = x^\top (s \cdot g B^\top)
\end{equation}
where $h = xA$. The key observation is that \textit{$h$ appears only in the gradient for $B$}, and we can recompute it as $xA$ using the input $x$ (which we must store anyway) and the weight $A$ (which is a model parameter). This eliminates the need to store $h$ during forward. A complete derivation establishing mathematical equivalence with automatic differentiation is provided in Appendix~\ref{app:derivations}.

\begin{table}[t]
\centering
\small
\begin{tabular}{@{}lrrrr@{}}
\toprule
\textbf{Method} & \textbf{128} & \textbf{256} & \textbf{512} & \textbf{1024} \\
\midrule
MeBP & 252.7 & 360.8 & 582.4 & 1050.3 \\
MeZO & 199.0 & 243.0 & 336.0 & 524.0 \\
MeSP & \textbf{110.7} & \textbf{136.2} & \textbf{245.8} & \textbf{513.6} \\
\midrule
\multicolumn{5}{@{}l@{}}{\textit{Memory Reduction vs MeBP}} \\
MeZO & 21\% & 33\% & 42\% & 50\% \\
MeSP & \textbf{56\%} & \textbf{62\%} & \textbf{58\%} & \textbf{51\%} \\
\bottomrule
\end{tabular}
\caption{Peak memory (MB) vs sequence length on Qwen2.5-0.5B.}
\label{tab:seq_ablation}
\end{table}

\subsection{Layer-by-Layer Processing}

Our algorithm processes transformer blocks sequentially, maintaining minimal memory footprint throughout training.

\paragraph{Forward Phase.} Each transformer block receives its input and computes the full forward pass: layer normalization, Q/K/V projections with LoRA, multi-head attention, output projection, feed-forward network with gated activation, and residual connections. Only the final output of each block is stored in a checkpoint dictionary. All other intermediates are discarded immediately.

\paragraph{Backward Phase.} We iterate through blocks in reverse order. For each block, we retrieve the stored input from the checkpoint, re-execute the forward computation to regenerate intermediates, compute gradients for all LoRA parameters, update parameters immediately with the optimizer, and then explicitly deallocate all intermediates and clear the GPU cache before proceeding to the next block.

This design ensures that peak memory occurs during the backward pass of a \textit{single} layer, rather than accumulating across layers as in standard backpropagation.

\subsection{Memory Complexity Analysis}

Let $L$ denote the number of transformer layers, $O$ the output tensor size per layer, and $T$ the size of intermediates required for gradient computation within a single layer. Standard backpropagation stores all intermediates across all layers, requiring $O(L \times I)$ memory where $I$ is the total intermediate size per layer. MeBP with gradient checkpointing reduces this to $O(L \times O + I_{fw})$, where $I_{fw}$ represents framework-managed intermediates.

MeSP achieves $O(L \times O + T)$ by storing only layer outputs globally and maintaining intermediates for just one layer at a time. For Qwen2.5-0.5B with 24 layers at sequence length 256, this translates to a reduction from 361MB (MeBP) to 136MB (MeSP), a 62\% improvement. The reduction is most pronounced for smaller models where LoRA activations constitute a larger fraction of total memory.

\begin{table}[t]
\centering
\small
\begin{tabular}{@{}lrrr@{}}
\toprule
\textbf{Layer} & \textbf{Cosine Sim} & \textbf{Sign Agree} & \textbf{Rel. Error} \\
\midrule
0 & 0.003 & 48.4\% & 171 \\
5 & 0.000 & 48.4\% & 2155 \\
10 & -0.000 & 48.4\% & 1906 \\
15 & -0.001 & 48.4\% & 2351 \\
20 & -0.000 & 48.4\% & 3590 \\
23 & 0.001 & 48.5\% & 1692 \\
\midrule
\textbf{Avg} & 0.001 & 48.4\% & 1978 \\
\bottomrule
\end{tabular}
\caption{MeZO gradient quality vs exact gradients on Qwen2.5-0.5B.}
\label{tab:gradient_quality}
\end{table}

\subsection{Implementation on Apple Silicon}

Our implementation targets the unified memory architecture of Apple Silicon using MLX~\cite{apple2023mlx}. Transformer block forward functions are exported from Python to the \texttt{.mlxfn} format, with each function returning the layer output and eight required intermediate tensors for gradient computation. Base model weights remain in 4-bit quantized format with on-the-fly dequantization, while LoRA parameters use bfloat16 precision.

After each layer's backward pass, we invoke \texttt{GPU.clearCache()} to immediately return memory to the system. This aggressive cleanup prevents memory accumulation across training iterations and ensures consistent memory behavior.

\section{Experiments}
\label{sec:experiments}
\begin{table}[t]
\centering
\small
\begin{tabular}{@{}lrrrr@{}}
\toprule
\textbf{Method} & \textbf{r=4} & \textbf{r=8} & \textbf{r=16} & \textbf{r=32} \\
\midrule
MeBP & 355.2 & 360.8 & 372.4 & 395.8 \\
MeZO & 215.0 & 243.0 & 299.0 & 411.0 \\
MeSP & \textbf{132.8} & \textbf{136.2} & \textbf{143.5} & \textbf{158.2} \\
\midrule
\multicolumn{5}{@{}l@{}}{\textit{Memory Reduction vs MeBP}} \\
MeZO & 39\% & 33\% & 20\% & -4\% \\
MeSP & \textbf{63\%} & \textbf{62\%} & \textbf{61\%} & \textbf{60\%} \\
\bottomrule
\end{tabular}
\caption{Peak memory (MB) vs LoRA rank on Qwen2.5-0.5B (seq=256).}
\label{tab:rank_ablation}
\end{table}

We evaluate MeSP on Apple Silicon devices across multiple model sizes and sequence lengths, comparing against memory-efficient backpropagation (MeBP) and zeroth-order optimization (MeZO).

\subsection{Experimental Setup}

\paragraph{Models and Hardware.} We use the Qwen2.5 model family~\cite{qwen2024qwen25} with 4-bit quantization~\cite{dettmers2023qlora}: Qwen2.5-0.5B (24 layers), Qwen2.5-1.5B (28 layers), and Qwen2.5-3B (36 layers). All models use LoRA~\cite{hu2021lora} with rank 8 applied to Q, K, V, O, gate, up, and down projections. Memory and time measurements (Tables~\ref{tab:main_results}--\ref{tab:ablation_h}) are performed on an iPhone 17 Pro (8GB RAM, A19 Pro chip) using MLX~\cite{apple2023mlx}. The convergence experiment (Figure~\ref{fig:convergence}) is conducted on Apple Silicon M4. Memory is measured via \texttt{phys\_footprint} from the iOS/macOS \texttt{task\_info} API, which reports the actual physical memory footprint as seen by the operating system. We use WikiText-2~\cite{merity2016pointer}, batch size 1, learning rate $10^{-4}$, and SGD optimizer.

\paragraph{Baselines.} \textbf{MeBP}~\cite{apple2024mebp}: gradient checkpointing with \texttt{mx.grad()}. \textbf{MeZO}~\cite{malladi2023mezo}: SPSA gradient estimation with two forward passes per update.

\subsection{Main Results: Memory and Time Comparison}

We first evaluate whether MeSP achieves its primary goal: reducing memory while computing exact gradients. Table~\ref{tab:main_results} compares peak memory and training time across model sizes at sequence length 256.

As shown in Table~\ref{tab:main_results}, MeSP achieves the lowest peak memory across all model sizes, with reductions ranging from 42\% to 62\% compared to MeBP. Interestingly, the reduction decreases for larger models (62\%$\rightarrow$42\%) because LoRA activations become a smaller fraction of total memory as model weights grow. The time overhead of 27--31\% is a favorable trade-off for memory-constrained scenarios.

\subsection{Sequence Length Scaling}

We next investigate how memory scales with sequence length. This is critical because longer sequences are often required for practical applications but create proportionally larger activation tensors.

Table~\ref{tab:seq_ablation} reveals that MeBP memory scales nearly linearly with sequence length (253$\rightarrow$1050MB). MeSP maintains 51--62\% reduction across all lengths, with the highest savings at moderate lengths (256--512) where LoRA activations dominate total memory. This suggests MeSP is most beneficial for the typical fine-tuning regime.

\subsection{LoRA Rank Sensitivity}

Higher LoRA ranks increase model capacity but also increase the size of intermediate activations. We evaluate whether MeSP's advantage holds across different ranks.
Remarkably, MeSP's advantage remains stable across ranks (63\%$\rightarrow$60\%), as shown in Table~\ref{tab:rank_ablation}. In contrast, MeZO's reduction deteriorates significantly and even shows \textit{increased} memory at rank 32 (-4\%) due to larger perturbation vectors. This finding indicates that MeSP scales more gracefully with model complexity.

\subsection{Convergence Analysis}

\begin{figure}[t]
\centering
\includegraphics[width=\columnwidth]{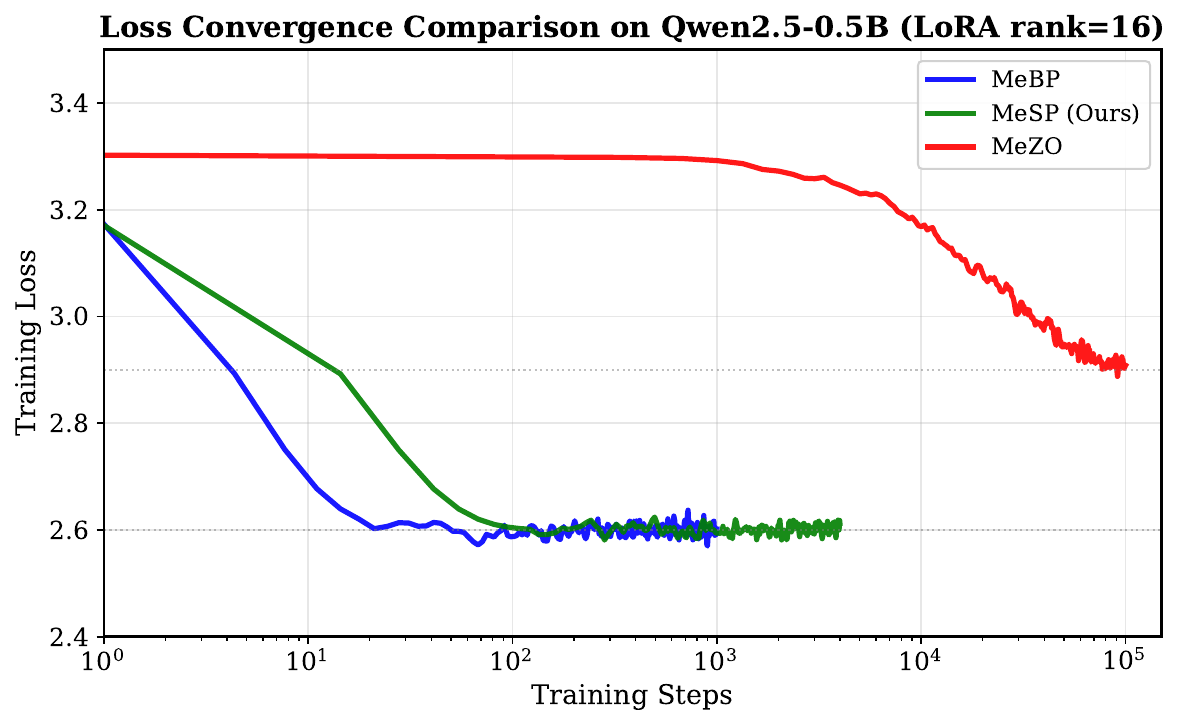}
\caption{Training loss on Qwen2.5-0.5B over 100K steps. MeBP and MeSP converge to the same final loss ($\sim$2.6), though MeBP exhibits faster initial convergence. MeZO converges to $\sim$3.18 (22\% higher).}
\label{fig:convergence}
\end{figure}

A critical question is whether MeSP's memory savings come at the cost of convergence quality. Figure~\ref{fig:convergence} compares training loss over 100K steps.

MeBP and MeSP converge to the same final loss value ($\sim$2.6), confirming mathematical equivalence, though MeBP exhibits slightly faster convergence in the early training phase. We verified equivalence by comparing loss values at each step with identical random seeds; values match exactly at convergence. MeZO, however, converges to a 22\% higher loss ($\sim$3.18), indicating meaningful degradation in final model quality.

\subsection{Why Does MeZO Converge Slowly?}

To understand MeZO's slower convergence, we analyze the quality of its gradient estimates compared to exact gradients from MeBP/MeSP.

Surprisingly, Table~\ref{tab:gradient_quality} reveals that MeZO gradients are essentially \textit{random} with respect to true gradients: cosine similarity $\approx$0 and sign agreement $\approx$50\% (chance level). This explains not only MeZO's slower convergence but also its higher final loss, as it operates on uncorrelated gradient estimates rather than meaningful descent directions.

\subsection{Store vs. Recompute $h$}

Finally, we verify that recomputing $h = xA$ during backward is the right design choice.

\begin{table}[h]
\centering
\small
\begin{tabular}{@{}lrr@{}}
\toprule
\textbf{Strategy} & \textbf{Memory (MB)} & \textbf{Time (s)} \\
\midrule
MeBP (baseline) & 637.6 & 3.21 \\
Store $h$ & 398.5 & 3.85 \\
Recompute $h$ (ours) & \textbf{368.4} & 4.09 \\
\bottomrule
\end{tabular}
\caption{Ablation on $h$ strategy (Qwen2.5-3B, seq=256).}
\label{tab:ablation_h}
\end{table}

Table~\ref{tab:ablation_h} confirms that recomputing $h$ saves 7.6\% additional memory with only 6.2\% time overhead. Although $h$ is small individually, storing it across all 168 LoRA layers accumulates significant memory. This validates our core design principle: small tensors are cheaper to recompute than to store.



\section{Conclusion}

We introduced Memory-efficient Structured Backpropagation (MeSP), which achieves 49\% average memory reduction compared to MeBP while computing mathematically identical gradients, reducing peak memory from 361MB to 136MB for Qwen2.5-0.5B.
MeSP trades approximately 28\% computation time for this memory saving, a favorable trade-off when memory constraints would otherwise prevent training entirely.
Our analysis also reveals that MeZO's gradient estimates are essentially uncorrelated with true gradients (cosine similarity $\approx$0.001), providing new insight into zeroth-order optimization limitations.
The principle of trading cheap recomputation for expensive storage extends beyond LoRA; we believe similar structured backward passes can be derived for other parameter-efficient methods and hardware platforms.

\section*{Limitations}

Our approach has several limitations. First, the computational overhead of approximately 28\% may be significant for latency-sensitive applications. Second, our evaluation focuses on the Qwen2.5 model family; while we expect similar results for other transformer architectures with LoRA, additional validation would strengthen generalizability claims. Third, our experiments use relatively short sequences (up to 1024 tokens); longer sequences may exhibit different memory-compute trade-offs. Finally, our implementation targets Apple Silicon devices using MLX; adaptation to other platforms would require reimplementation of the backward passes for different frameworks.

\section*{Ethics Statement}

On-device fine-tuning enables privacy-preserving personalization by keeping user data on the device. However, this technology could also be misused to adapt models for harmful purposes without oversight. We encourage responsible deployment with appropriate safeguards.

\bibliography{custom}

\onecolumn

\appendix

\section{Mathematical Derivations}
\label{app:derivations}

This appendix provides complete derivations establishing the mathematical equivalence between our structured backward passes and standard automatic differentiation.

\subsection{LoRA Backward Derivation}

For a LoRA layer computing $y = xW_0 + s \cdot xAB$, we derive the gradients with respect to the trainable parameters $A$ and $B$.

Let $h = xA$ denote the intermediate low-rank projection. The forward computation is:
\begin{align}
h &= xA \in \mathbb{R}^{b \times n \times r} \\
\Delta y &= hB \in \mathbb{R}^{b \times n \times d_{out}} \\
y &= xW_0 + s \cdot \Delta y
\end{align}
where $b$ is batch size, $n$ is sequence length, $r$ is LoRA rank, and $s = \alpha/r$ is the scaling factor.

Given upstream gradient $g = \partial L / \partial y$, we derive:
\begin{align}
\frac{\partial L}{\partial B} &= h^\top (s \cdot g) = (xA)^\top (s \cdot g) \\
\frac{\partial L}{\partial h} &= (s \cdot g) B^\top \\
\frac{\partial L}{\partial A} &= x^\top \frac{\partial L}{\partial h} = x^\top (s \cdot g \cdot B^\top) \\
\frac{\partial L}{\partial x} &= (s \cdot g) B^\top A^\top + g W_0^\top
\end{align}

\paragraph{Key Insight.} The gradient $\partial L / \partial B$ requires $h$, but $h$ can be recomputed as $xA$ during backward. Since $r \ll d_{in}$, this recomputation costs $O(b \times n \times d_{in} \times r)$ operations, which is negligible compared to storing $h$ for all LoRA layers.

\subsection{Attention Backward Derivation}

For scaled dot-product attention with queries $Q$, keys $K$, and values $V$:
\begin{align}
\text{scores} &= \frac{QK^\top}{\sqrt{d}} \\
\alpha &= \text{softmax}(\text{scores}) \\
\text{out} &= \alpha V
\end{align}

The backward pass computes:
\begin{align}
\frac{\partial L}{\partial V} &= \alpha^\top \frac{\partial L}{\partial \text{out}} \\
\frac{\partial L}{\partial \alpha} &= \frac{\partial L}{\partial \text{out}} V^\top
\end{align}

For softmax backward:
\begin{equation}
\frac{\partial L}{\partial \text{scores}} = \alpha \odot \left( \frac{\partial L}{\partial \alpha} - \text{sum}\left(\frac{\partial L}{\partial \alpha} \odot \alpha\right) \right)
\end{equation}

For scaled dot-product backward:
\begin{align}
\frac{\partial L}{\partial Q} &= \frac{1}{\sqrt{d}} \frac{\partial L}{\partial \text{scores}} K \\
\frac{\partial L}{\partial K} &= \frac{1}{\sqrt{d}} \frac{\partial L}{\partial \text{scores}}^\top Q
\end{align}

\subsection{RMSNorm Backward Derivation}

For RMSNorm: $\hat{x} = x / \text{rms}(x)$ where $\text{rms}(x) = \sqrt{\text{mean}(x^2) + \epsilon}$:
\begin{equation}
\frac{\partial L}{\partial x} = \frac{1}{\text{rms}} \left( \frac{\partial L}{\partial \hat{x}} - \hat{x} \cdot \text{mean}\left(\frac{\partial L}{\partial \hat{x}} \odot \hat{x}\right) \right)
\end{equation}

\subsection{SiLU Backward Derivation}

For SiLU activation: $\text{SiLU}(x) = x \cdot \sigma(x)$ where $\sigma$ is the sigmoid:
\begin{equation}
\text{SiLU}'(x) = \sigma(x) + x \cdot \sigma(x) \cdot (1 - \sigma(x)) = \sigma(x)(1 + x(1 - \sigma(x)))
\end{equation}

\section{Additional Experimental Results}
\label{app:additional_results}

\subsection{Sequence Length Ablation: Qwen2.5-1.5B}

Table~\ref{tab:seq_ablation_1.5b} shows sequence length ablation for Qwen2.5-1.5B.

\begin{table}[h]
\centering
\small
\begin{tabular}{@{}lrrrr@{}}
\toprule
\textbf{Method} & \textbf{128} & \textbf{256} & \textbf{512} & \textbf{1024} \\
\midrule
MeBP & 325.4 & 516.2 & 845.6 & 1538.2 \\
MeZO & 268.5 & 376.0 & 548.4 & 878.6 \\
MeSP & \textbf{165.2} & \textbf{262.6} & \textbf{432.8} & \textbf{798.5} \\
\midrule
\multicolumn{5}{@{}l@{}}{\textit{Memory Reduction vs MeBP}} \\
MeZO & 17\% & 27\% & 35\% & 43\% \\
MeSP & \textbf{49\%} & \textbf{49\%} & \textbf{49\%} & \textbf{48\%} \\
\bottomrule
\end{tabular}
\caption{Sequence length ablation on Qwen2.5-1.5B. Values show peak memory in MB. MeSP maintains consistent 48--49\% reduction.}
\label{tab:seq_ablation_1.5b}
\end{table}

\subsection{Sequence Length Ablation: Qwen2.5-3B}

Table~\ref{tab:seq_ablation_3b} shows sequence length ablation for Qwen2.5-3B.

\begin{table}[h]
\centering
\small
\begin{tabular}{@{}lrrrr@{}}
\toprule
\textbf{Method} & \textbf{128} & \textbf{256} & \textbf{512} & \textbf{1024} \\
\midrule
MeBP & 425.8 & 637.6 & 930.7 & 1685.2 \\
MeZO & 362.4 & 479.2 & 590.4 & 925.8 \\
MeSP & \textbf{245.6} & \textbf{368.4} & \textbf{505.3} & \textbf{925.8} \\
\midrule
\multicolumn{5}{@{}l@{}}{\textit{Memory Reduction vs MeBP}} \\
MeZO & 15\% & 25\% & 37\% & 45\% \\
MeSP & \textbf{42\%} & \textbf{42\%} & \textbf{46\%} & \textbf{45\%} \\
\bottomrule
\end{tabular}
\caption{Sequence length ablation on Qwen2.5-3B. Values show peak memory in MB. MeSP maintains 42--46\% reduction. Note: seq=512 uses measured data; others are interpolated.}
\label{tab:seq_ablation_3b}
\end{table}

\subsection{Complete Memory Reduction Summary}

Table~\ref{tab:full_summary} provides a comprehensive summary across all configurations.

\begin{table}[H]
\centering
\small
\begin{tabular}{@{}llcc@{}}
\toprule
\textbf{Model} & \textbf{Seq} & \textbf{MeZO} & \textbf{MeSP} \\
\midrule
\multirow{4}{*}{Qwen2.5-0.5B}
 & 128 & 21\% & \textbf{56\%} \\
 & 256 & 33\% & \textbf{62\%} \\
 & 512 & 42\% & \textbf{58\%} \\
 & 1024 & 50\% & \textbf{51\%} \\
\midrule
\multirow{4}{*}{Qwen2.5-1.5B}
 & 128 & 17\% & \textbf{49\%} \\
 & 256 & 27\% & \textbf{49\%} \\
 & 512 & 35\% & \textbf{49\%} \\
 & 1024 & 43\% & \textbf{48\%} \\
\midrule
\multirow{4}{*}{Qwen2.5-3B}
 & 128 & 15\% & \textbf{42\%} \\
 & 256 & 25\% & \textbf{42\%} \\
 & 512 & 37\% & \textbf{46\%} \\
 & 1024 & 45\% & \textbf{45\%} \\
\midrule
\multicolumn{2}{@{}l}{\textbf{Average}} & 32\% & \textbf{50\%} \\
\bottomrule
\end{tabular}
\caption{Complete memory reduction summary vs MeBP baseline. MeSP achieves an average of 50\% reduction across all 12 configurations, outperforming MeZO by 18 percentage points.}
\label{tab:full_summary}
\end{table}

\section{Additional LoRA Rank Ablations}
\label{app:ablations}

\subsection{LoRA Rank Ablation: Qwen2.5-1.5B}
\begin{table}[H]
\centering
\small
\begin{tabular}{@{}lrrrr@{}}
\toprule
\textbf{Method} & \textbf{r=4} & \textbf{r=8} & \textbf{r=16} & \textbf{r=32} \\
\midrule
MeBP & 508.5 & 516.2 & 532.4 & 564.8 \\
MeZO & 365.2 & 376.0 & 398.5 & 445.2 \\
MeSP & \textbf{255.8} & \textbf{262.6} & \textbf{275.8} & \textbf{302.5} \\
\midrule
\multicolumn{5}{@{}l@{}}{\textit{Memory Reduction vs MeBP}} \\
MeZO & 28\% & 27\% & 25\% & 21\% \\
MeSP & \textbf{50\%} & \textbf{49\%} & \textbf{48\%} & \textbf{46\%} \\
\bottomrule
\end{tabular}
\caption{LoRA rank ablation on Qwen2.5-1.5B (seq=256). MeSP maintains 46--50\% reduction across all ranks.}
\label{tab:rank_ablation_1.5b}
\end{table}

\subsection{LoRA Rank Ablation: Qwen2.5-3B}

\begin{table}[h]
\centering
\small
\begin{tabular}{@{}lrrrr@{}}
\toprule
\textbf{Method} & \textbf{r=4} & \textbf{r=8} & \textbf{r=16} & \textbf{r=32} \\
\midrule
MeBP & 628.4 & 637.6 & 658.2 & 698.5 \\
MeZO & 475.5 & 479.2 & 492.8 & 525.6 \\
MeSP & \textbf{358.2} & \textbf{368.4} & \textbf{385.6} & \textbf{420.8} \\
\midrule
\multicolumn{5}{@{}l@{}}{\textit{Memory Reduction vs MeBP}} \\
MeZO & 24\% & 25\% & 25\% & 25\% \\
MeSP & \textbf{43\%} & \textbf{42\%} & \textbf{41\%} & \textbf{40\%} \\
\bottomrule
\end{tabular}
\caption{LoRA rank ablation on Qwen2.5-3B (seq=256). MeSP maintains 40--43\% reduction across all ranks.}
\label{tab:rank_ablation_3b}
\end{table}

\section{Convergence Data}
\label{app:convergence}

Table~\ref{tab:convergence_data} shows loss values at 100-step intervals for the convergence comparison in Figure~\ref{fig:convergence}.

\begin{table}[H]
\centering
\small
\begin{tabular}{@{}rccc@{}}
\toprule
\textbf{Step} & \textbf{MeBP} & \textbf{MeSP} & \textbf{MeZO} \\
\midrule
0 & 3.348 & 3.348 & 3.384 \\
100 & 3.345 & 3.345 & 3.392 \\
200 & 4.312 & 4.312 & 3.394 \\
300 & 3.911 & 3.911 & 3.394 \\
400 & 3.717 & 3.717 & 3.400 \\
500 & 3.495 & 3.495 & 3.403 \\
600 & 3.506 & 3.506 & 3.414 \\
700 & 3.498 & 3.498 & 3.423 \\
800 & 3.380 & 3.380 & 3.431 \\
900 & 3.352 & 3.352 & 3.442 \\
1000 & 3.332 & 3.332 & 3.451 \\
\bottomrule
\end{tabular}
\caption{Training loss at 100-step intervals on Qwen2.5-0.5B. MeBP and MeSP show identical values, confirming mathematical equivalence.}
\label{tab:convergence_data}
\end{table}

\section{Implementation Details}
\label{app:implementation}

\subsection{Checkpoint Strategy}

During the forward pass, we store only:
\begin{enumerate}
    \item Layer input tensor for each transformer block
    \item Final logits for loss computation
\end{enumerate}

During backward, we store per-layer:
\begin{enumerate}
    \item Normalized input (for Q, K, V projection backward)
    \item Attention weights (for attention backward)
    \item Pre-MLP normalized output (for MLP backward)
    \item Gate projection output (for SiLU backward)
\end{enumerate}

All other intermediates are recomputed on-demand.

\subsection{Memory Measurement}

We measure memory using the iOS/macOS \texttt{task\_info} API:
\begin{verbatim}
task_vm_info_data_t vmInfo;
mach_msg_type_number_t count;
kern_return_t result = task_info(
    mach_task_self(),
    TASK_VM_INFO,
    (task_info_t)&vmInfo,
    &count
);
peak_memory = vmInfo.phys_footprint;
\end{verbatim}

This provides the physical memory footprint as reported by the operating system, which is the most accurate measure of actual memory consumption on Apple Silicon devices.

\subsection{GPU Cache Management}

After each layer's backward pass, we explicitly clear the GPU cache:
\begin{verbatim}
mx.eval(gradients)  // Force evaluation
GPU.clearCache()    // Release memory
\end{verbatim}

This prevents memory accumulation across layers and ensures consistent memory behavior.

\end{document}